\documentclass[11pt,a4paper]{article}
\usepackage{acl}
\usepackage{times}
\usepackage{xspace}
\usepackage{amsmath}
\usepackage{latexsym}
\usepackage{threeparttable}
\usepackage{multirow}
\usepackage{pbox}
\usepackage{pgfplots}
\usepackage{paralist}
\usepackage{ragged2e,array,booktabs}
\usepackage{graphicx}
\usepackage{subcaption}
\usepackage{amsfonts}
\usepackage{relsize}

\usetikzlibrary{positioning}
\usetikzlibrary{calc}
\usetikzlibrary{fit}
\usetikzlibrary{decorations.text}
\usetikzlibrary{shapes.multipart}

\usepackage{microtype}

\usepackage{xspace}
\usepackage{threeparttable}
\usepackage{multirow}
\usepackage{graphicx}
\usepackage{tabularx}
\usepackage{mathtools}
\usepackage{amsmath}
\usepackage{booktabs}
\usepackage{multicol}
\usepackage{multirow}
\usepackage{dsfont}
\usepackage{amsmath}

\DeclareMathOperator{\sign}{sign}
\DeclareMathOperator{\argmin}{argmin}
\DeclareMathOperator{\argmax}{argmax}

\newcommand{\figref}[2][]{Figure#1~\ref{#2}\xspace}
\newcommand{\tabref}[2][]{Table#1~\ref{#2}\xspace}
\newcommand{\secref}[1]{Section~\ref{#1}\xspace}
\newcommand{\customeqref}[2][]{Equation#1~\ref{#2}\xspace}
\newcommand{\appendixref}[1]{Appendix~\ref{#1}\xspace}

\newcommand{\class}[1]{\textsc{#1}\xspace}
\newcommand{\aae}{\class{AAE}}
\newcommand{\sae}{\class{SAE}}
\newcommand{\happy}{\class{happy}}
\newcommand{\sad}{\class{sad}}

\newcommand{\dataset}[1]{\textbf{#1}\xspace}

\newcommand{\moji}{\dataset{Moji}}
\newcommand{\bios}{\dataset{Bios}}

\newcommand{\bert}{BERT\xspace}
\newcommand{\deepmoji}{DeepMoji\xspace}

\newcommand{\method}[1]{\textsf{#1}\xspace}
\newcommand{\vanilla}{\method{CE}}
\newcommand{\diffadv}{\method{Adv}}
\newcommand{\inlp}{\method{INLP}}

\newcommand{\difference}{\method{EO$_\mathrm{{CLA}}$}}
\newcommand{\mean}{\method{EO$_\mathrm{{GLB}}$}}
\newcommand{\differencemax}{\method{EO$_\mathrm{{CLA}}^{\mathrm{max}}$}}
\newcommand{\differencemin}{\method{EO$_\mathrm{{CLA}}^{\mathrm{min}}$}}
\newcommand{\meanmax}{\method{EO$_\mathrm{{GLB}}^{\mathrm{max}}$}}
\newcommand{\meanmin}{\method{EO$_\mathrm{{GLB}}^{\mathrm{min}}$}}
\newcommand{\ds}{\method{DS}}
\newcommand{\rw}{\method{RW}}

\newcommand{\fairbatch}{\method{FairBatch}}
\newcommand{\constrained}{\method{Constrained}}

\newcommand{\gap}{GAP\xspace}

\newcommand{\microF}{F$_{1}^{\text{micro}}$\xspace}
\newcommand{\macroF}{F$_{1}^{\text{macro}}$\xspace}

\newcommand{\z}{\phantom{0}}

\newcommand{\ssig}{\ensuremath{\dagger}}

\title{Optimising Equal Opportunity Fairness in Model Training}

\author{Aili Shen\thanks{\ ~Equal contributors to this work.}$^\spadesuit$\quad
        Xudong Han\footnotemark[1]$^\spadesuit$ \quad
    	Trevor Cohn$^\spadesuit$\quad
    	Timothy Baldwin$^{\spadesuit\heartsuit}$\quad
    	Lea Frermann$^\spadesuit$ \\
        $\spadesuit$ School of Computing and Information Systems\\
        The University of Melbourne \\
        Victoria 3010, Australia\\
        $\heartsuit$ Department of Natural Language Processing, MBZUAI\\
	    \url{{aili.shen,t.cohn,tbaldwin,lfrermann}@unimelb.edu.au}, \url{xudongh1@student.unimelb.edu.au}
  }

\date{}

\begin{document}

\maketitle

\begin{abstract}
Real-world datasets often encode stereotypes and societal biases. Such biases can be implicitly captured by trained models, leading to biased predictions and exacerbating existing societal preconceptions. Existing debiasing methods, such as adversarial training and removing protected information from representations, have been shown to reduce bias. However, a disconnect between fairness criteria and training objectives makes it difficult to reason theoretically about the effectiveness of different techniques. In this work, we propose two novel training objectives which directly optimise for the widely-used criterion of {\it equal opportunity}, and show that they are effective in reducing bias while maintaining high performance over two classification tasks.
\end{abstract}

\section{Introduction and Background}
\label{introduction}

Modern neural machine learning has achieved great success across a range of classification tasks. 
However, when applied over real-world data, especially in high-stakes settings such as hiring processes and loan approvals, care must be taken to assess the fairness of models. This is because real-world datasets generally encode societal preconceptions and stereotypes, thereby leading to models trained on such datasets to amplify existing bias and make biased predictions (i.e., models perform unequally towards different subgroups of individuals). This kind of unfairness has been reported over various NLP tasks, such as part-of-speech tagging \cite{Hovy:15,Li:18,Han:21b}, sentiment analysis \cite{Blodgett:16,Shen:21}, and image activity recognition \cite{Wang:19,Zhao:17}. 

Various methods have been proposed to mitigate bias, including adversarial training, and pre- and post-processing strategies. Adversarial training aims to make it difficult for a discriminator to predict protected attribute values from learned representations \cite{Han:21,Elazar:18,Madras:18}. Pre- and post-processing strategies vary greatly in approach, including  transforming the original dataset to reduce protected attribute discrimination while retaining dataset utility \citep{Calmon:17}, iteratively removing protected attribute information from (fixed) learned representations \citep{Shauli:20}, or reducing bias amplification by injecting corpus-level constraints during inference \citep{Zhao:17}.


However, training strategies and optimisation objectives are generally disconnected from fairness metrics which directly measure the extent to which different groups are treated (in)equitably. This makes it difficult to understand the effectiveness of previous debiasing methods from a theoretical perspective. In this work, we propose to explicitly incorporate equal opportunity into our training objective, thereby achieving bias reduction. This paper makes the following contributions: 
\begin{compactenum}
	\item We are the first to propose a weighted training objective that directly implements fairness metrics.
	\item Observing that model performance for different classes can vary greatly, we further propose a variant of our method, taking both bias reduction among protected attribute groups and bias reduction among different classes into consideration.
	\item Experimental results over two tasks show that both proposed methods are effective at achieving fairer predictions, while maintaining performance. 
\end{compactenum}
Our code is available at: \url{https://github.com/AiliAili/Difference_Mean_Fair_Models}.

\section{Related Work}

\subsection{Fairness Criteria}

Various criteria have been proposed to capture different types of discrimination, such as group fairness \cite{Hardt:16,Zafar:17,Cho:20,Zhao:20}, individual fairness \cite{Saeed:19,Yurochkin:20,Dwork:12}, and causality-based fairness \cite{Wu:19,Zhang:18,Zhang:18b}. In this work, we focus on group fairness, whereby a model should perform equally across different demographic subgroups. 

To quantify how predictions vary across different demographic subgroups, demographic parity \cite{Feldman:15,Zafar:17b,Cho:20}, equal opportunity \cite{Hardt:16,Madras:18}, and equalised odds \cite{Cho:20,Hardt:16,Madras:18} are widely used to measure fairness. \textit{Demographic parity} ensures that models achieve the same positive rate for each demographic subgroup, oblivious the ground-truth target label. \textit{Equal opportunity} requires that a model achieves the same true positive rate (TPR) across different subgroups, considering only instances with a positive label. \textit{Equalised odds} goes one step further in requiring not only the same TPR but also the same false positive rate (FPR) across groups. 

Demographic parity, equal opportunity, and equalised odds only focus on the prediction outcome for one specific target label (i.e.\ a ``positive'' class) in a binary classification setting, but does not apply fairness directly to multi-class settings, 
when fairness for different subgroups across all classes is required. Equal opportunity can be generalised by extending the ``positive'' class to each target class, as we do in our work.

\subsection{Debiasing Methods}
\label{sec:debiaising}

A broad range of methods has been proposed to learn fair models. Based on where debiasing occurs, in terms of dataset processing, model training, and inference, we follow \citet{Cho:20} in categorising methods into: (1) pre-processing, (2) post-processing, and (3) in-processing. 

\textbf{Pre-processing methods} manipulate the original dataset to mitigate discrimination \cite{Wang:19,Xu:18,Feldman:15,Calmon:17,De-Arteaga:19}. For example, \citet{Calmon:17} transform the original dataset to reduce discrimination while retaining dataset utility. Class imbalance methods used in bias reduction, such as dataset sampling \cite{Kubat:97,Wallace:11}, instance reweighting \cite{Cui:19,Li:20,Lin:17}, and weighted max-margin \cite{Cao:19}, also belong to this category. For example, \citet{Lahoti:20}, \citet{Subramanian:21}, and \citet{Han:21c} reweight instances by taking the (inverse of) joint distribution of the protected attribute classes and main task classes into consideration. 
\citet{Wang:19} and \citet{Han:21c} down-sample the majority protected attribute group within each target class, and train on the resulting balanced dataset.

\textbf{Post-processing methods} calibrate the prediction outcome or learned representations of models to achieve fair predictions \cite{Hardt:16,Pleiss:17,Zhao:17,Shauli:20}. For example, \citet{Zhao:17} enforce a corpus-level constraint during inference to reduce bias. \citet{Shauli:20} iteratively remove protected attribute information from representations generated by an fixed encoder, by iteratively training a discriminator over the projected attribute and projecting the representation into the discriminator's null space. 

\textbf{In-processing methods} learn fair models during model training. One family of approaches is based on constrained optimisation, incorporating fairness measures as regularisation terms or constraints \cite{Zafar:17b,Subramanian:21b,Donini:18,Narasimhan:18,Cho:20}. 
For example, \citet{Zafar:17} translate equalised odds into constraints on FPR and FNR across groups, and solve using constraint programming. \citet{Cho:20} adopt kernel density estimation to quantify demographic parity and equalised odds, but in a manner which is limited to low-dimensional data and binary classification tasks.
Another line of work is to use adversarial training to obtain fair models, in jointly training an encoder and discriminator(s) over the encoded representations such that the discriminator(s) are ineffective at predicting the protected attribute values from learned representations \cite{Han:21,Elazar:18,Madras:18,Zhanghu:18,Agarwal:18,Roh:20}.
Elsewhere, \citet{Shen:21} use contrastive learning to learn fair models by simultaneously pushing instances belonging to the same target class closer and pulling instances belonging to the same protected attribute class further apart. 

The most relevant work to ours is \emph{FairBatch} \cite{Roh:21}. It proposes to formulate the original task as a bi-level optimisation problem, where the inner optimiser is the standard training algorithm and the outer optimiser is responsible for adaptively adjusting the sampling probabilities of instances with a given target class and protected attribute value, based on the equal opportunity metric achieved by the intermediate inner model. That is, they adaptively adjust the instance \textit{resampling} probability during training to reduce bias. However, different from {FairBatch}, whose resampling strategy is bound by the sampling probability $[0,1]$, our proposed method achieves bias reduction by \textit{reweighting} instances during training, where the reweighting range is unbounded, leading to greater flexibility in trading off performance and fairness.

\section{Methodology}

\subsection{Preliminaries}

Suppose we have some data $X\in\mathds{R}^{n}$, target labels $Y \in C$, 
and protected attribute values $A=\{0, 1\}$, where $C$ is the number of target classes for a given task. 

\paragraph{Equal opportunity} A classifier is said to satisfy equal opportunity if its prediction is conditionally independent of the protected attribute $A$ given the target label $Y$, \{$P(\hat{y}=y|Y=y,A=0)=P(\hat{y}=y|Y=y,A=1)$\} for $\forall y \in Y$. Here, $\hat{y}$ is a prediction outcome, $y \in Y$ and $a \in A$. As mentioned above, we slightly modify the definition of equal opportunity by allowing $y$ to be each candidate target class, accommodating multi-class settings. We explicitly address the fairness criterion across {\it all} target classes by promoting comparable true positive rates across protected classes.


\subsection{Optimising Equal Opportunity}

Instead of using a fairness proxy \cite{Zafar:17b} or kernel density estimation to quantify fairness \cite{Cho:20}, we propose to optimise equal opportunity by directly minimising the absolute difference in loss between different subsets of instances belonging to the same target label but with different protected attribute classes,
\begin{equation}
\begin{aligned}
\mathcal{L}_{\text{eo}}^{\text{class}}
= \mathcal{L}_{ce}+\lambda\sum_{y=\in C}\sum_{a \in A}|\mathcal{L}^{y,a}_{ce}-\mathcal{L}_{ce}^{y}|
\end{aligned}
\label{difference}
\end{equation}
Here, $\mathcal{L}_{ce}$ denotes the average cross-entropy loss based on instances in the batch; 
$\mathcal{L}_{ce}^{y,a}$ denotes the average cross-entropy loss computed over instances with the target label $y$ and the protected attribute label~$a$; and $\mathcal{L}_{ce}^{y}$ denotes the average cross-entropy loss computed over all instances with target label~$y$.
Our proposed loss $\mathcal{L}_{\text{eo}}^{\text{class}}$ is the weighted sum of the overall cross-entropy and the sum of the cross-entropy difference for each target label overall and that conditioned on the target label, thereby capturing both performance and fairness. This method is denoted as \difference, as it captures \underline{cla}ss-wise equal opportunity.

\subsection{Equal Opportunity across Classes}

One drawback of \difference is that it only focuses on optimising equal opportunity, 
ignoring the fact that the performance for different classes can vary greatly, especially when the dataset is skewed. To learn fair models not only towards demographic subgroups but also across target classes, we propose a variant of \customeqref{difference}, by introducing one additional constraint on top of equal opportunity to encourage the label-wise cross entropy loss terms to align. Formally: $\mathcal{L}_{ce}^{y_{1}}\approx \mathcal{L}_{ce}^{y_{2}}$, where $y_{1}\neq y_{2}$, and $y_{1} \in Y$, $y_{2} \in Y$. This objective encourages equal opportunity not only for demographic subgroups but also across different target classes:
\begin{equation}
\mathcal{L}_{\text{eo}}^{\text{global}}=\mathcal{L}_{ce}+\lambda\sum_{y=\in C}\sum_{a \in A}|\mathcal{L}^{y, a}_{ce}-\mathcal{L}_{ce}|
\label{mean}
\end{equation}
This method is denoted as \mean, short for \underline{gl}o\underline{b}al equal opportunity.

\subsection{Theory}

In this section, we show how our training objective is related to equal opportunity in the binary classification and binary protected attribute setting. Note that our proof naturally extends to cases where the numbers of target classes and/or protected attribute values are greater than two as described in \customeqref[s]{difference} and \ref{mean}.  

Let $m_{y,a}$ be the number of training instances with target label $y$ and protected attribute $a$ in a batch. For example, $m_{1,0}$ denotes the number of instances with target label $1$ and protected attribute $0$ in the batch. 
Let $\mathcal{L}^{y,a}$ be the average loss for instances with target label $y$ and protected attribute $a$. For example, $\mathcal{L}^{1,0}$ is the average loss for instances with target label $1$ and protected attribute $0$.

\subsubsection{Cross-Entropy Loss}

The vanilla cross-entropy loss is computed as:
\begin{equation}
  \small
  \frac{1}{N}(m_{0,0}\mathcal{L}^{0,0}+m_{0,1}\mathcal{L}^{0,1}+m_{1,0}\mathcal{L}^{1,0}+m_{1,1}\mathcal{L}^{1,1})
\label{ce_proof}
\end{equation}
which is the average loss over different subsets of instances with a given target label and protected attribute class.

\subsubsection{Difference Loss}

The \difference method defined in \customeqref{difference} can be written as:
\begin{equation}
\begin{aligned}
\mathcal{L}_{\text{eo}}^{\text{class}}&= \mathcal{L}_{ce}+\lambda\sum_{y,a}|\mathcal{L}^{y,a}_{ce}-\mathcal{L}_{ce}^{y}|\\
&= \sum_{y,a}\Big[\frac{m_{y,a}}{N}\mathcal{L}_{ce}^{y,a} + \lambda \sign_{y,a}{(\mathcal{L}^{y,a}_{ce}-\mathcal{L}_{ce}^{y})} \Big],\\
&= \sum_{y,a}\Big[(\frac{m_{y,a}}{N}+\lambda \sign_{y,a})\mathcal{L}_{ce}^{y,a} \\
&\qquad \qquad -\lambda \sign_{y,a}{\mathcal{L}_{ce}^{y}} \Big],
\label{difference_proof_1}
\end{aligned}
\end{equation}
where $\sign$ is a sign function, and $\sign_{y,a} = \sign{(\mathcal{L}^{y,a}_{ce}-\mathcal{L}_{ce}^{y})}$. 
Noting that for binary protected attributes, $\sign_{y,a} = -\sign_{y,\neg a}$, and $\sum_{y,a}\mathcal{L}_{ce}^{y}=0,\forall y$ in this case:
\begin{equation}
\begin{aligned}
\mathcal{L}_{\text{eo}}^{\text{class}}&= \sum_{y,a} (\frac{m_{y,a}}{N}+\lambda \sign_{y,a})\mathcal{L}_{ce}^{y,a} 
\label{difference_proof}
\end{aligned}
\end{equation}

By comparing \customeqref[s]{ce_proof} and \ref{difference_proof}, we can see that for target label $y$, our method dynamically increases the weight for poorly-performing subsets (i.e.\ $\sign_{y,g} = 1$) by $\lambda$, and decreases the weight for well-performing subsets ($\sign_{y,g} = -1$) by $\lambda$, thereby leading to fairer predictions by adjusting the weight for instances with different protected attribute classes conditioned on a given target label. 

\subsubsection{From Binary Cross-Entropy to True Positive Rate} 
Using the definition of binary cross-entropy
\begin{equation*}
-[y_{i}\cdot \log(p(y_{i})) + (1-y_{i})\cdot \log(1-p(y_{i}))],
\end{equation*}
the loss for a certain subset, e.g., the subset of instances with target label $1$ and protected attribute class $0$, can be simplified as:
\begin{equation}
\begin{aligned}
\mathcal{L}^{1,0} = &-\frac{1}{m_{1,0}}\sum^{m_{1,0}}_{j=1}\Big( y_{j}\cdot \log(p(\hat{y}_{j})) \Big.\\
&\Big. \qquad+(1-y_{j})\cdot \log(1-p(\hat{y}_{j})) \Big)\\
&=-\frac{1}{m_{1,0}}\sum^{m_{1,0}}_{j=1} \log(p(\hat{y}_{j}))
\end{aligned}
\end{equation}
Notice that $p(\hat{y}_{j})$ is equivalent to $p(\hat{y}_{j} = 1)$, making $\mathcal{L}^{1,0} =-\frac{1}{m_{1,0}}\sum^{m_{1,0}}_{j=1} \log(p(\hat{y}_{j}))$ an unbiased estimator of $-\log p(\hat{y}=1|y=1)$, which approximates $- \log$ TPR.

Minimising the expectation of the absolute difference between $\mathcal{L}^{1,0}$ and $\mathcal{L}^{1,1}$ can approximate the true positive rate difference between two groups with the same target label $1$:
\begin{equation*}
\begin{aligned}
&\argmin_{\theta} \mathbb{E}(|\mathcal{L}^{1,0}-\mathcal{L}^{1,1}|)\\
&= \argmin |-\log p(\hat{y}=1|y=1,g=0)\\
&\qquad-(-\log p(\hat{y}=1|y=1,g=1))|\\
&\approx \argmin |\log \frac{TPR_{1,0}}{TPR_{1,1}}| \\
&= \argmin |TPR_{1,0} - TPR_{1,1}|
\label{tpr_diff}
\end{aligned}
\end{equation*}
This demonstrates that minimising the absolute difference between $\mathcal{L}^{1,0}$ and $\mathcal{L}^{1,1}$ is roughly equivalent to minimising the TPR difference between two groups with the same target label, which is precisely the formulation of equal opportunity, as described in \secref{evaluation_metrics}. Therefore, the second term in our proposed method (\customeqref{difference}) is optimising directly for equal opportunity.

\section{Experiments}
Our experiments compare the performance and fairness of our methods against various competitive baselines, and across two classification tasks.

\subsection{Baselines}

We compare our proposed methods \difference and \mean against the following seven baselines:

\begin{compactenum}
	\item \vanilla: train the model with cross-entropy loss and no explicit bias mitigation. 
	\item \inlp: first train the model with cross-entropy loss to obtain dense representations, and iteratively apply null-space projection to the learned representations to remove protected attribute information \cite{Shauli:20}. The resulting representations are used to make predictions. 
	\item \diffadv: jointly train the model with cross-entropy loss and an ensemble of three adversarial discriminators for the projected attribute, with an orthogonality constraint applied to the discriminators to encourage diversity \cite{Han:21}. 
	\item \ds: downsample the dataset corresponding to the protected attribute conditioned on a given target label \cite{Han:21c}.
	\item \rw: reweight instances based on the (inverse) joint distribution of the protected attribute classes and target classes \cite{Han:21c}. 
	\item \constrained: formulate the task as a constrained optimisation problem, where equal opportunity is incorporated as constraints \cite{Subramanian:21b}.
	\item \fairbatch: formulate the model training as a bi-level optimisation problem, as described in \secref{sec:debiaising} \cite{Roh:21}.
\end{compactenum}

\subsection{Experiment Setup}

For each task, we first obtain document representations from their corresponding pretrained models, which are not finetuned during training. Then document representations are fed into two fully-connected layers with a hidden size of 300d. For all experiments, we use the {A}dam optimiser \cite{Kingma:15} to optimise the model for at most 60 epochs with early stopping and a patience of~5. 
All models are trained and evaluated on the same dataset splits, and models are selected based on their performance on the development set. We finetune the learning rate, batch size, and extra hyperparameters introduced by the corresponding debiasing methods for each model on each dataset (see the Appendix for details). 
Noting the complexity of model selection given the multi-objective accuracy--fairness tradeoff and the absence of a standardised method for selecting models based on both criteria in fairness research, we determine the best-achievable accuracy for a given model, and select the hyperparameter settings that reduce bias while maintaining accuracy as close as possible to the best-achievable value (all based on the dev set). We leave the development of a fair and robust model selection method to future work.

\subsection{Evaluation Metrics}
\label{evaluation_metrics}

To evaluate the performance of models on the main task, we adopt \microF and \macroF for all our datasets, taking class imbalance into consideration, especially in the multi-class setting.

To evaluate fairness, we follow previous work~\cite{De-Arteaga:19,Shauli:20} and adopt root mean square TPR gap over all classes, which is defined as
\begin{equation*}
\text{\gap}=\sqrt{\frac{1}{|C|}\sum_{y\in{Y}}(\mathrm{GAP}^{\mathrm{TPR}}_{y}})^{2},
\end{equation*}
where $\mathrm{GAP}^{\mathrm{TPR}}_{y}=|\mathrm{TPR}_{y,a}-\mathrm{TPR}_{y, \neg a}|, ~~y\in{Y}$, and $\mathrm{TPR}_{a,y}=\mathds{P}(\hat{y}=y|y,a)$, indicating the proportion of correct predictions among instances with target label $y$ and protected attribute label $a$. $\mathrm{GAP}^{\mathrm{TPR}}_{y}$ measures the absolute performance difference between demographic subgroups conditioned on target label $y$, and a value of $0$ indicates that the model makes predictions independent of the protected attribute. 

\subsection{Twitter Sentiment Analysis}

\subsubsection{Task and Dataset}
\label{moji_task}

For our first dataset, the task is to predict the binary sentiment for a given English tweet, where each tweet is also annotated with a binary protected attribute indirectly capturing the ethnicity of the tweet author as either African American English (\aae) or Standard American English (\sae). 
Following previous studies \cite{Shauli:20,Han:21,Shen:21}, we adopt the dataset of \citet{Blodgett:16} (\moji hereafter), where the training dataset is balanced with respect to both sentiment and ethnicity but skewed in terms of sentiment--ethnicity combinations (40\% \happy-\aae, 10\% \happy-\sae, 10\% \sad-\aae, and 40\% \sad-\sae, respectively). The number of instances in the training, dev, and test sets are 100K, 8K, and 8K, respectively. The dev and test set are balanced in terms of sentiment--ethnicity combinations.

\subsubsection{Implementation Details}

Following previous work \cite{Elazar:18,Shauli:20,Han:21}, we use \deepmoji \cite{Felbo:17}, a model pretrained over 1.2 billion English tweets, as the encoder to obtain text representations. The parameters of \deepmoji are fixed in training. Hyperparameter settings are provided in \appendixref{hyperparameter_moji}.

\subsubsection{Experimental Results}

\tabref{moji_result} presents the results over the \moji test set. Compared to \vanilla, \inlp and \diffadv moderately reduce model bias while simultaneously improving model performance. Surprisingly, both \ds and \rw reduce \gap substantially and achieve the joint best \microF, indicating that the biased prediction is mainly due to the imbalanced distribution of protected attribute classes conditioned on a given target label, and the imbalanced distribution of sentiment--ethnicity combinations.\footnote{However, it does not hold the other way around as demonstrated by previous studies \cite{Wang:19}, indicating that a balanced dataset either in terms of target label and protected attribute combination, or in terms of protected attribute class distribution conditioned on target classes, can still lead to biased predictions.} However, the drawback of dataset imbalance methods is that they lack the flexibility to control the performance--fairness tradeoff. Both \constrained and \fairbatch also effectively reduce bias and achieve improved performance. 
Both of our methods, \difference and \mean, achieve competitive performance on the main task with the largest bias reduction. For all models except \inlp, we can see that incorporating debiasing techniques leads to improved performance on the main task. We hypothesise that incorporating debiasing techniques (either in the form of adversarial training, data imbalance methods, or optimising towards equal opportunity) acts as a form of regularisation, thereby reducing the learned correlation between the protected attribute and main task label, and encouraging models to learn task-specific representations.

\begin{table}[!t]  
	\centering   	
	\scalebox{1}{ 
		\begin{threeparttable}
			\begin{tabular}{lcc}  
				\toprule  
				Model &\microF$\uparrow$ &\gap$\downarrow$	 \\ \midrule	
				\vanilla &72.09$\pm$0.65  &40.21$\pm$1.23   \\				
				\inlp &72.81$\pm$0.01  &36.81$\pm$3.49   	\\						
				\diffadv  &74.47$\pm$0.68 &30.59$\pm$2.94    \\
				\ds &76.16$\pm$0.28 &14.96$\pm$1.08  \\
				\rw &\z\textbf{76.21}$\pm$0.16$^{\ssig}$ &14.70$\pm$0.86  \\
				\constrained &75.22$\pm$0.20 &15.92$\pm$4.86 \\\vspace{1mm}
				\fairbatch &75.81$\pm$0.17 &15.36$\pm$3.07  \\
				\difference  &75.03$\pm$0.25  &\z\textbf{10.83}$\pm$1.40$^{\ssig}$      \\ 	
				\mean &75.20$\pm$0.20  &11.49$\pm$1.07      \\ \bottomrule
			\end{tabular}
	\end{threeparttable}}
	\caption{Experimental results on the \moji test set (averaged over 10 runs); \textbf{Bold} $=$ Best Performance; $\uparrow =$ the higher the better; $\downarrow =$ the lower the better. The best result is marked with ``\ssig'' if the difference over the next-best method is statistically significant (based on a one-tailed Wilcoxon signed-rank test; $p<0.05$), noting that if the best method is one of our methods, we compare it to the next-best method which is not our own.}
	\label{moji_result}
\end{table}

\begin{figure}[t!]
	\centering
	\includegraphics[width=\linewidth]{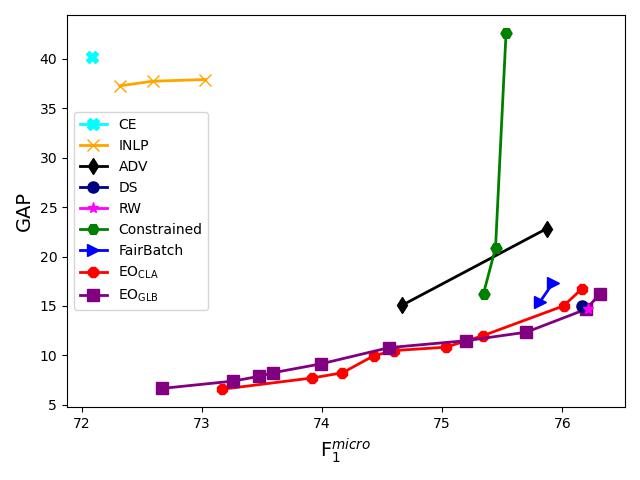}
	\caption{\microF vs. \gap of different models on the \moji test set, as we vary the most sensitive hyperparameter for each model.}
	\label{moji_pareto}
\end{figure}

\paragraph{Performance--Fairness tradeoff.} We plot the tradeoff between \microF and \gap for all models on the \moji test set in \figref{moji_pareto}. In this, we vary the most-sensitive hyperparameter for each model: the number of iterations for \inlp, the $\lambda$ weight for adversarial loss for \diffadv, the step size of adjusting resampling probability for \fairbatch, and the weight for minimising the loss difference for \difference and \mean.\footnote{For \vanilla, \ds, and \rw, there is no hyperparameter that controls the tradeoff between model performance and bias reduction.} As we can see, \inlp has limited capacity to reduce bias, and the performance for the main task is slightly worse than the other methods. Compared with \diffadv, \constrained, and \fairbatch, our proposed methods \difference and \mean achieve fairer predictions while maintaining competitive performance (bottom right). Another advantage of our methods is that they allow for greater variability in the performance--fairness tradeoff, demonstrating the effectiveness and superiority of our proposed method. Note that only the pareto points for each model are plotted. For example, for \diffadv, we experimented with 7 values of $\lambda$, but the results are captured by only two pareto points. 

\subsection{Profession Classification}

\subsubsection{Task and Dataset}

For our second dataset, the task is to predict a person's occupation given their biography \cite{De-Arteaga:19}, where each short online biography is labelled with one of 28 occupations (main task label) and binary gender (protected attribute). Following previous work \cite{De-Arteaga:19,Shauli:20}, the number of instances in the training, dev, and test sets are 257K, 40K, and 99K, respectively.\footnote{There are slight differences between our dataset and that used by previous studies \cite{De-Arteaga:19,Shauli:20} as a small number of biographies were no longer available on the web when we crawled them.}

\subsubsection{Implementation Details}
Following the work of \citet{Shauli:20}, we use the ``CLS'' token representation of the pretrained uncased \bert-base \cite{Devlin:19} to obtain text representations, and keep \bert fixed during training. Hyperparameter settings for all models are provided in \appendixref{hyperparameter_bios}.


\begin{table}[!t]  
	\centering   	
	\scalebox{0.80}{ 
		\begin{threeparttable}
			\begin{tabular}{lccc}  
				\toprule  
				Model &\macroF$\uparrow$ &\microF$\uparrow$ &\gap$\downarrow$	 \\   \midrule
				\vanilla &\z\textbf{75.95}$\pm$0.10$^{\ssig}$  &\z\textbf{82.19}$\pm$0.04 $^{\ssig}$ &16.68$\pm$0.46   \\
				\inlp   &71.44$\pm$0.40 &79.54$\pm$0.18  &13.52$\pm$1.54   	\\		
				\diffadv  &70.88$\pm$2.31  &79.72$\pm$1.02 &16.78$\pm$0.87  \\
				\ds &67.73$\pm$0.26   &78.48$\pm$0.10  &\z9.17$\pm$0.41  \\ 
				\rw &69.21$\pm$0.36  &76.18$\pm$0.32 &\z\textbf{8.58}$\pm$0.49$^{\ssig}$    \\
				\fairbatch  &75.14$\pm$0.28  &81.82$\pm$0.07  &10.80$\pm$1.04  \vspace{1mm} \\ 	
				\difference &72.07$\pm$0.18  &81.52$\pm$0.06  &12.80$\pm$0.42     \\
				\mean &75.11$\pm$0.18  &81.74$\pm$0.07 &12.72$\pm$0.51     \\	
				\bottomrule
			\end{tabular}
	\end{threeparttable}}
	\caption{Experimental results on the \bios test set (averaged over 10 runs). The best result is marked with ``\ssig'' if the difference over the next-best method is statistically significant (based on a one-tailed Wilcoxon signed-rank test; $p<0.05$), noting that if the best method is one of our methods, we compare it to the next-best method which is not our own.
	}
	\label{bios_result}
\end{table}

\subsubsection{Experimental Results}

\tabref{bios_result} shows the results on the \bios test set.\footnote{We omit results for \constrained as it did not converge on this data set, presumably because of its brittleness over multi-class classification tasks.} We can see that \diffadv is unable to reduce \gap even at the cost of performance in terms of \microF and \macroF. Both \ds and \rw reduce bias in terms of \gap, at the cost of a drop in performance, in terms of \microF and \macroF. We attribute this to the dramatic decrease in the number of training instances for \ds, and the myopia of \rw in only taking the ratio of occupation--gender combinations into consideration but not the difficulty of each target class. Among \inlp, \fairbatch, \difference, and \mean, we can see that \fairbatch achieves a reasonable bias reduction with the least performance drop. This is due to it dynamically adjusting the resampling probability during training. Comparing \difference and \mean, we can see that \mean is better able to deal with the dataset class imbalance (reflected in \macroF), while reducing bias.

\begin{figure}[t!]
	\centering
	\includegraphics[width=\linewidth]{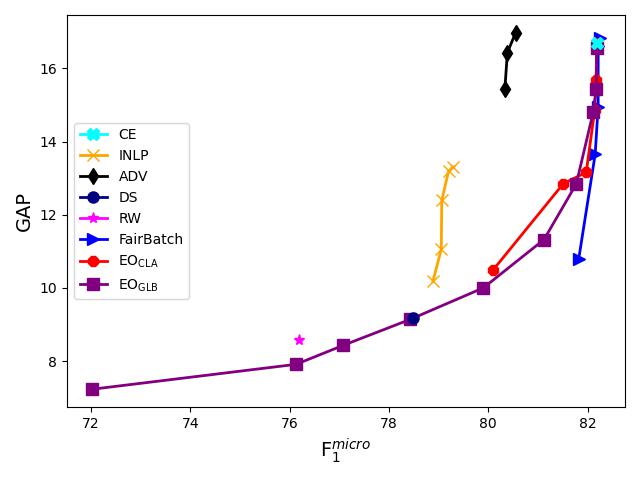}
	\caption{\microF vs. \gap of different models on the \bios test set, as we vary the most sensitive hyperparameter for each model.}
	\label{micro_pareto}
\end{figure}

\begin{figure}[t!]
	\centering
	\includegraphics[width=\linewidth]{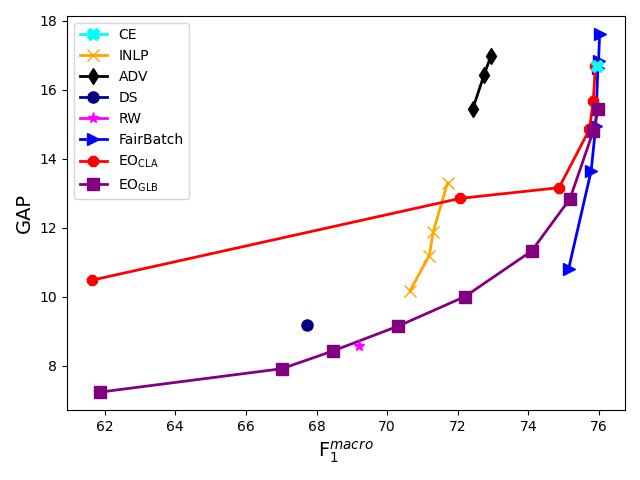}
	\caption{\macroF vs. \gap of different models on the \bios test set, as we vary the most sensitive hyperparameter for each model.}
	\label{macro_pareto}
\end{figure}

\paragraph{Performance--Fairness tradeoff.} 
\figref{micro_pareto} shows the \microF--\gap tradeoff plot for the \bios test set. We can see that \inlp and \diffadv reduce bias at the cost of performance, as do \ds and \rw. Compared with \fairbatch, \difference and \mean provide greater control in terms of performance--fairness tradeoff, such as achieving a smaller \gap with a slight decrease of \microF. 
A similar trend is also observed for the  \macroF--\gap tradeoff as shown in \figref{macro_pareto}. Although \difference is outperformed by \fairbatch, \mean provides greater control in terms of performance--fairness tradeoff, suggesting an advantage of \mean in enforcing fairness across target classes, especially for the imbalanced dataset.



\section{Analysis}

To better understand the effectiveness of our proposed methods, we perform two sets of experiments: (1) an ablation study, and (2) an analysis of training efficiency. 

\subsection{Ablation Study}

\difference can be reformulated as $\mathcal{L}_{ce}+\lambda\sum_{y\in C}\{\max(\mathcal{L}^{y,a}_{ce}, \mathcal{L}^{y,\neg a}_{ce})-\min(\mathcal{L}^{y,a}_{ce}, \mathcal{L}^{y,\neg a}_{ce})\}$, effectively assigning more weight to worse-performing instances ($\argmax \text{loss}$) and less weight to better-performing instances ($\argmin \text{loss}$). To explore the impact of adjusting weights on model performance, we experiment with two versions: (1) $\mathcal{L}_{ce}+\lambda\sum_{y\in C}\max(\mathcal{L}^{y,a}_{ce}, \mathcal{L}^{y,\neg a}_{ce})$, denoted as \differencemax, where we assign higher weights to worse-performing instances without changing the weights assigned to better-performing instances; and (2) $\mathcal{L}_{ce}-\lambda\sum_{y\in C}\min(\mathcal{L}^{y,a}_{ce}, \mathcal{L}^{y,\neg a}_{ce})$, denoted as \differencemin, where we assign smaller weights to better-performing instances without changing the weights assigned to worse-performing instances. Correspondingly, for \mean, we have \meanmax and \meanmin. Hyperparameter settings for each model can be found in \appendixref{ablation_parameters}.

\tabref[s]{ablation_result_moji} and \ref{ablation_result_bios} show the results for the different models on \moji and \bios. 
We can see that the full \difference and \mean both achieve better bias reduction than ablated {\it min} and {\it max} counterparts on \moji, while maintaining similar levels of performance in terms of \microF.\footnote{For the max and min versions of both \difference and \mean, we finetune with the corresponding best-performing $\lambda$, respectively. A smaller \gap value cannot be achieved by further adjusting/increasing the value of $\lambda$.} On \bios, we can see that \differencemax outperforms \difference in bias reduction and model performance except for \microF, indicating that it is beneficial for bias reduction to increase the weight for worse-performing instances. On the other hand, \differencemin is inferior to \difference in terms of both bias reduction and performance. We conjecture that reducing the weights for better-performing instances is harmful for model performance (especially for minority classes) over datasets with imbalanced distributions, as is the case for \bios.\footnote{This is in line with previous research \cite{Swayamdipta:20}, which shows that easy-to-learn instances are important in optimising models.} Among the three variants of \mean, \meanmax slightly improves performance on the main task and maintains the same level of bias reduction as \mean, while \meanmin improves performance on the main task but does not reduce bias. Overall, these results show that our two methods perform best in their original formulations.

\begin{table}[!t]  
	\centering   	
	\scalebox{0.9}{ 
		\begin{threeparttable}
			\begin{tabular}{lcc}  
				\toprule  
				Model &\microF$\uparrow$ &\gap$\downarrow$ \\   \midrule
				\difference  &75.03$\pm$0.25  &10.83$\pm$1.40     \\ 	
				\differencemax &75.92$\pm$0.10   &13.79$\pm$1.64    \\ 
				\differencemin &75.33$\pm$0.19   &14.50$\pm$1.78  \\ \midrule
				\mean &75.20$\pm$0.20  &11.49$\pm$1.07 \\
				\meanmax &76.31$\pm$0.10 &16.47$\pm$0.90 \\
				\meanmin &76.27$\pm$0.13 &18.01$\pm$0.40 \\
				\bottomrule
			\end{tabular}
	\end{threeparttable}}
	\caption{Ablation results over \moji test set (averaged over 10 runs).}
	\label{ablation_result_moji}
\end{table}

\begin{table}[!t]  
	\centering   	
	\scalebox{0.9}{ 
		\begin{threeparttable}
			\begin{tabular}{lccc}  
				\toprule  
				Model &\macroF$\uparrow$ &\microF &\gap$\downarrow$ \\   \midrule 
				\difference &72.07$\pm$0.18  &81.52$\pm$0.06  &12.80$\pm$0.42   \\
				\differencemax &72.09$\pm$0.19 &79.95$\pm$0.12 &8.98$\pm$0.43   \\ 
				\differencemin &53.17$\pm$0.53 &76.66$\pm$0.23 &19.22$\pm$1.68   \\ \midrule
				\mean &75.11$\pm$0.18  &81.74$\pm$0.07 &12.72$\pm$0.51   \\
				\meanmax &75.37$\pm$0.06 &81.89$\pm$0.03  &12.47$\pm$0.51   \\ 
				\meanmin &75.95$\pm$0.12  &82.19$\pm$0.05 &16.74$\pm$0.42  \\ \bottomrule
			\end{tabular}
	\end{threeparttable}}
	\caption{Ablation results over \bios test set (averaged over 10 runs).}
	\label{ablation_result_bios}
\end{table}

\begin{figure}[t!]
	\centering
	\includegraphics[width=\linewidth]{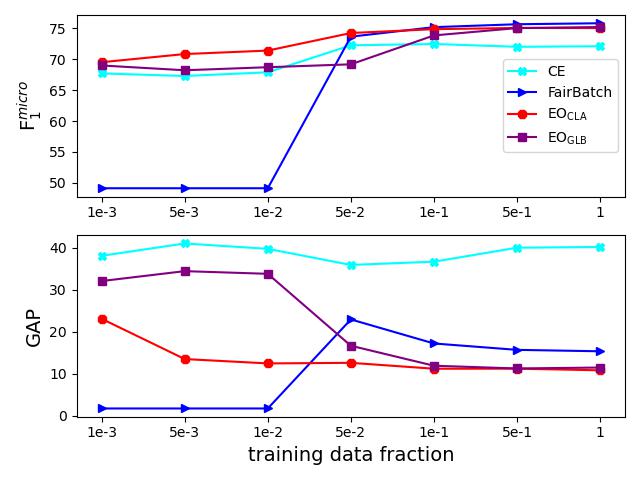}
	\caption{\microF vs.\ \gap of different models on the \moji test set. The full training set is 100K instances.}
	\label{moji_data_size}
\end{figure}

%

\begin{figure}[t!]
	\centering
	\includegraphics[width=\linewidth]{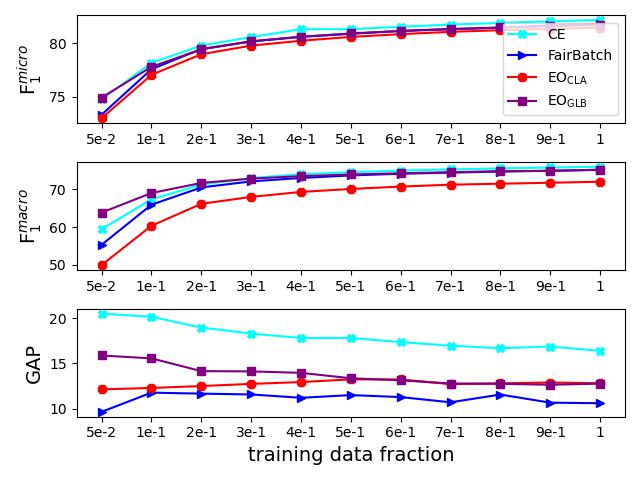}
	\caption{\microF and \macroF vs.\ \gap of different models on the \bios test set. The full training set is 257K.}
	\label{bios_micro_macro_data_size}
\end{figure}

\subsection{Training Efficiency}
\label{training_efficiency}

To understand the training efficiency of the different models, we perform experiments with varying training data sizes on both \moji and \bios. Based on results from \tabref[s]{moji_result} and \ref{bios_result}, we provide results for \vanilla, \fairbatch, \difference, and \mean. 

\figref{moji_data_size} presents the results for \moji. When the proportion of training data is no larger than 1K, \fairbatch is unable to learn a decent model, while both \difference and \mean are still effective. As we increase the number of training instances, improved performance on the main task can been observed for all models, and larger bias reduction is achieved for all models except \vanilla. Overall, \difference and \mean perform well in low-resource settings and achieve better bias reduction for larger volumes of training instances, demonstrating their superiority. 

\begin{figure}[t!]
	\centering
	\includegraphics[width=\linewidth]{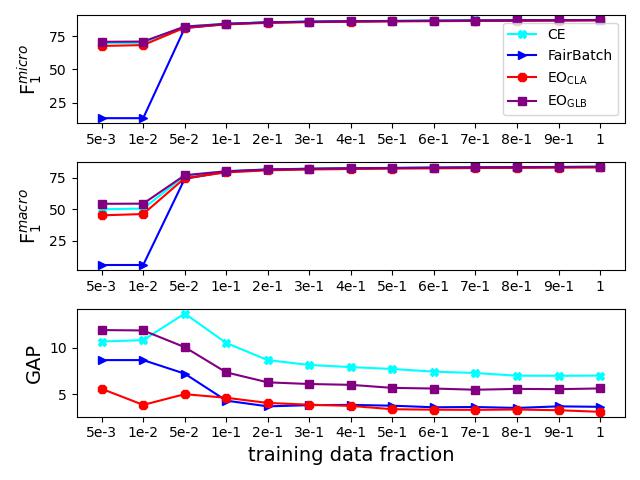}
	\caption{\microF, \macroF, vs. \gap of different models on the subset of \bios data. Here, instances are from the top-$8$ most common classes, whose proportion is greater than $4$\% in the original dataset, resulting into a full training dataset size of 188K.}
	\label{bios_micro_macro_data_size_8_classes}
\end{figure}

\figref{bios_micro_macro_data_size} presents the results for \bios. We see that \fairbatch outperforms \difference and \mean, especially in terms of \macroF and \gap. Our explanation is that \fairbatch adopts a resampling strategy, while our method adopts a reweighting strategy. Although statistically equivalent, resampling outperforms reweighting when combined with stochastic gradient algorithms \cite{An:21}. The data imbalance in \bios exacerbates this effect. To verify this, we generated a version of \bios with only instances belonging to the top-$8$ most common classes, whose ratio in the original training set is bigger than 4\%. 
\figref{bios_micro_macro_data_size_8_classes} presents results with the subset of dataset consisting of the top-$8$ most common classes. The plots show a similar trend as observed for the \moji dataset on this relatively balanced dataset. Specifically, when the training dataset is small, \fairbatch is unable to learn a decent model, while both \difference and \mean are still effective. 
 
\subsection{Limitations}
\label{limitations}
Consistent with previous work, we did not finetune the underlying pretrained models in obtaining document representations in this work.  Finetuning may further remove biases encoded in the pretrained models, which we leave to future work. This work focused only on datasets with binary protected attributes, and future experiments should explore the methods' generalization to higher-arity attributes. For both \inlp and \diffadv, we follow experimental setup from the original papers, noting that the \texttt{fairlib}~\citep{Han:22} debiasing framework\footnote{\url{https://pypi.org/project/fairlib/}} --- which was developed after this work was done --- recently showed that both models can obtain better performance and fairness scores with a larger budget for hyperparameter finetuning.

\section{Conclusion}

We proposed to incorporate fairness criteria into model training, in explicitly optimising for equal opportunity by minimising the loss difference over different subgroups conditioned on the target label. To deal with data imbalance based on the target-label, we proposed a variant of our method which promotes fairness across all target labels. Experimental results over Twitter sentiment analysis and profession classification tasks show the effectiveness and flexibility of our proposed methods. 

\section*{Ethical Considerations}
Our works aims to achieve fairer models, contributing to equal treatment for different demographic subgroups. However, its usage in the real world should be carefully calibrated/auditioned as debiasing for one projected attribute does not guarantee fairness for other protected attributes. In this work, due to the limitations of the dataset, we treat gender as binary, which is not perfectly aligned with the real world.

\section*{Acknowledgements}

We thank anonymous reviewers for their helpful feedback. This research was supported by The University of Melbourne’s Research Computing Services and the Petascale Campus Initiative. 
This work was funded in part by the Australian Research Council.

\bibliographystyle{acl_natbib}
\bibliography{naacl2022}

\begin{thebibliography}{49}
\expandafter\ifx\csname natexlab\endcsname\relax\def\natexlab#1{#1}\fi

\bibitem[{Agarwal et~al.(2018)Agarwal, Beygelzimer, Dud{\'{\i}}k, Langford, and
  Wallach}]{Agarwal:18}
Alekh Agarwal, Alina Beygelzimer, Miroslav Dud{\'{\i}}k, John Langford, and
  Hanna~M. Wallach. 2018.
\newblock A reductions approach to fair classification.
\newblock In \emph{Proceedings of the 35th International Conference on Machine
  Learning}, pages 60--69.

\bibitem[{An et~al.(2021)An, Ying, and Zhu}]{An:21}
Jing An, Lexing Ying, and Yuhua Zhu. 2021.
\newblock Why resampling outperforms reweighting for correcting sampling bias
  with stochastic gradients.
\newblock In \emph{Proceedings of the 9th International Conference on Learning
  Representations}.

\bibitem[{Blodgett et~al.(2016)Blodgett, Green, and O{'}Connor}]{Blodgett:16}
Su~Lin Blodgett, Lisa Green, and Brendan O{'}Connor. 2016.
\newblock Demographic dialectal variation in social media: {A} case study of
  {A}frican-{A}merican {E}nglish.
\newblock In \emph{Proceedings of the 2016 Conference on Empirical Methods in
  Natural Language Processing}, pages 1119--1130.

\bibitem[{Cao et~al.(2019)Cao, Wei, Gaidon, Ar{\'{e}}chiga, and Ma}]{Cao:19}
Kaidi Cao, Colin Wei, Adrien Gaidon, Nikos Ar{\'{e}}chiga, and Tengyu Ma. 2019.
\newblock Learning imbalanced datasets with label-distribution-aware margin
  loss.
\newblock In \emph{Advances in Neural Information Processing Systems}, pages
  1565--1576.

\bibitem[{Cho et~al.(2020)Cho, Hwang, and Suh}]{Cho:20}
Jaewoong Cho, Gyeongjo Hwang, and Changho Suh. 2020.
\newblock A fair classifier using kernel density estimation.
\newblock In \emph{Advances in Neural Information Processing Systems}.

\bibitem[{Cui et~al.(2019)Cui, Jia, Lin, Song, and Belongie}]{Cui:19}
Yin Cui, Menglin Jia, Tsung{-}Yi Lin, Yang Song, and Serge~J. Belongie. 2019.
\newblock Class-balanced loss based on effective number of samples.
\newblock In \emph{Proceedings of the {IEEE} Conference on Computer Vision and
  Pattern Recognition}, pages 9268--9277.

\bibitem[{De-Arteaga et~al.(2019)De-Arteaga, Romanov, Wallach, Chayes, Borgs,
  Chouldechova, Geyik, Kenthapadi, and Kalai}]{De-Arteaga:19}
Maria De-Arteaga, Alexey Romanov, Hanna Wallach, Jennifer Chayes, Christian
  Borgs, Alexandra Chouldechova, Sahin Geyik, Krishnaram Kenthapadi, and
  Adam~Tauman Kalai. 2019.
\newblock Bias in bios: {A} case study of semantic representation bias in a
  high-stakes setting.
\newblock In \emph{Proceedings of the Conference on Fairness, Accountability,
  and Transparency}, pages 120--128.

\bibitem[{Devlin et~al.(2019)Devlin, Chang, Lee, and Toutanova}]{Devlin:19}
Jacob Devlin, Ming-Wei Chang, Kenton Lee, and Kristina Toutanova. 2019.
\newblock {BERT}: {P}re-training of deep bidirectional transformers for
  language understanding.
\newblock In \emph{Proceedings of the 2019 Conference of the North {A}merican
  Chapter of the Association for Computational Linguistics: Human Language
  Technologies, Volume 1 (Long and Short Papers)}, pages 4171--4186.

\bibitem[{Donini et~al.(2018)Donini, Oneto, Ben{-}David, Shawe{-}Taylor, and
  Pontil}]{Donini:18}
Michele Donini, Luca Oneto, Shai Ben{-}David, John Shawe{-}Taylor, and
  Massimiliano Pontil. 2018.
\newblock Empirical risk minimization under fairness constraints.
\newblock In \emph{Advances in Neural Information Processing Systems}, pages
  2796--2806.

\bibitem[{Dwork et~al.(2012)Dwork, Hardt, Pitassi, Reingold, and
  Zemel}]{Dwork:12}
Cynthia Dwork, Moritz Hardt, Toniann Pitassi, Omer Reingold, and Richard~S.
  Zemel. 2012.
\newblock Fairness through awareness.
\newblock In \emph{Innovations in Theoretical Computer Science 2012}, pages
  214--226.

\bibitem[{Elazar and Goldberg(2018)}]{Elazar:18}
Yanai Elazar and Yoav Goldberg. 2018.
\newblock Adversarial removal of demographic attributes from text data.
\newblock In \emph{Proceedings of the 2018 Conference on Empirical Methods in
  Natural Language Processing}, pages 11--21.

\bibitem[{Felbo et~al.(2017)Felbo, Mislove, S{\o}gaard, Rahwan, and
  Lehmann}]{Felbo:17}
Bjarke Felbo, Alan Mislove, Anders S{\o}gaard, Iyad Rahwan, and Sune Lehmann.
  2017.
\newblock Using millions of emoji occurrences to learn any-domain
  representations for detecting sentiment, emotion and sarcasm.
\newblock In \emph{Proceedings of the 2017 Conference on Empirical Methods in
  Natural Language Processing}, pages 1615--1625.

\bibitem[{Feldman et~al.(2015)Feldman, Friedler, Moeller, Scheidegger, and
  Venkatasubramanian}]{Feldman:15}
Michael Feldman, Sorelle~A. Friedler, John Moeller, Carlos Scheidegger, and
  Suresh Venkatasubramanian. 2015.
\newblock Certifying and removing disparate impact.
\newblock In \emph{Proceedings of the 21th {ACM} {SIGKDD} International
  Conference on Knowledge Discovery and Data Mining}, pages 259--268.

\bibitem[{Han et~al.(2021{\natexlab{a}})Han, Baldwin, and Cohn}]{Han:21c}
Xudong Han, Timothy Baldwin, and Trevor Cohn. 2021{\natexlab{a}}.
\newblock Balancing out bias: {A}chieving fairness through training
  reweighting.
\newblock \emph{CoRR}, abs/2109.08253.

\bibitem[{Han et~al.(2021{\natexlab{b}})Han, Baldwin, and Cohn}]{Han:21b}
Xudong Han, Timothy Baldwin, and Trevor Cohn. 2021{\natexlab{b}}.
\newblock Decoupling adversarial training for fair {NLP}.
\newblock In \emph{Findings of the Association for Computational Linguistics},
  pages 471--477.

\bibitem[{Han et~al.(2021{\natexlab{c}})Han, Baldwin, and Cohn}]{Han:21}
Xudong Han, Timothy Baldwin, and Trevor Cohn. 2021{\natexlab{c}}.
\newblock Diverse adversaries for mitigating bias in training.
\newblock In \emph{Proceedings of the 16th Conference of the European Chapter
  of the Association for Computational Linguistics: Main Volume}, pages
  2760--2765.

\bibitem[{Han et~al.(2022)Han, Shen, Li, Cohn, Baldwin, and Frermann}]{Han:22}
Xudong Han, Aili Shen, Yitong Li, Trevor Cohn, Timothy Baldwin, and Lea
  Frermann. 2022.
\newblock fairlib: A unified framework for assessing and improving
  classification fairness.
\newblock \emph{arXiv preprint}.

\bibitem[{Hardt et~al.(2016)Hardt, Price, and Srebro}]{Hardt:16}
Moritz Hardt, Eric Price, and Nati Srebro. 2016.
\newblock Equality of opportunity in supervised learning.
\newblock In \emph{Advances in Neural Information Processing Systems}, pages
  3315--3323.

\bibitem[{Hovy and S{\o}gaard(2015)}]{Hovy:15}
Dirk Hovy and Anders S{\o}gaard. 2015.
\newblock Tagging performance correlates with author age.
\newblock In \emph{Proceedings of the 53rd annual meeting of the Association
  for Computational Linguistics and the 7th International Joint Conference on
  Natural Language Processing (volume 2: Short papers)}, pages 483--488.

\bibitem[{Kingma and Ba(2015)}]{Kingma:15}
Diederik~P. Kingma and Jimmy Ba. 2015.
\newblock Adam: {A} method for stochastic optimization.
\newblock In \emph{Proceedings of the 3rd International Conference on Learning
  Representations}.

\bibitem[{Kubat and Matwin(1997)}]{Kubat:97}
Miroslav Kubat and Stan Matwin. 1997.
\newblock Addressing the curse of imbalanced training sets: {One-Sided}
  selection.
\newblock In \emph{Proceedings of the Fourteenth International Conference on
  Machine Learning}, pages 179--186.

\bibitem[{Lahoti et~al.(2020)Lahoti, Beutel, Chen, Lee, Prost, Thain, Wang, and
  Chi}]{Lahoti:20}
Preethi Lahoti, Alex Beutel, Jilin Chen, Kang Lee, Flavien Prost, Nithum Thain,
  Xuezhi Wang, and Ed~Chi. 2020.
\newblock Fairness without demographics through adversarially reweighted
  learning.
\newblock In \emph{Advances in Neural Information Processing Systems}.

\bibitem[{Li et~al.(2020)Li, Sun, Meng, Liang, Wu, and Li}]{Li:20}
Xiaoya Li, Xiaofei Sun, Yuxian Meng, Junjun Liang, Fei Wu, and Jiwei Li. 2020.
\newblock Dice loss for data-imbalanced {NLP} tasks.
\newblock In \emph{Proceedings of the 58th Annual Meeting of the Association
  for Computational Linguistics}, pages 465--476.

\bibitem[{Li et~al.(2018)Li, Baldwin, and Cohn}]{Li:18}
Yitong Li, Timothy Baldwin, and Trevor Cohn. 2018.
\newblock Towards robust and privacy-preserving text representations.
\newblock In \emph{Proceedings of the 56th Annual Meeting of the Association
  for Computational Linguistics}, pages 25--30.

\bibitem[{Lin et~al.(2017)Lin, Goyal, Girshick, He, and Doll{\'a}r}]{Lin:17}
Tsung-Yi Lin, Priya Goyal, Ross Girshick, Kaiming He, and Piotr Doll{\'a}r.
  2017.
\newblock Focal loss for dense object detection.
\newblock In \emph{Proceedings of the IEEE international conference on computer
  vision}, pages 2980--2988.

\bibitem[{Madras et~al.(2018)Madras, Creager, Pitassi, and Zemel}]{Madras:18}
David Madras, Elliot Creager, Toniann Pitassi, and Richard~S. Zemel. 2018.
\newblock Learning adversarially fair and transferable representations.
\newblock In \emph{Proceedings of the 35th International Conference on Machine
  Learning,}, pages 3381--3390.

\bibitem[{Narasimhan(2018)}]{Narasimhan:18}
Harikrishna Narasimhan. 2018.
\newblock Learning with complex loss functions and constraints.
\newblock In \emph{International Conference on Artificial Intelligence and
  Statistics}, pages 1646--1654.

\bibitem[{du~Pin~Calmon et~al.(2017)du~Pin~Calmon, Wei, Vinzamuri, Ramamurthy,
  and Varshney}]{Calmon:17}
Fl{\'{a}}vio du~Pin~Calmon, Dennis Wei, Bhanukiran Vinzamuri,
  Karthikeyan~Natesan Ramamurthy, and Kush~R. Varshney. 2017.
\newblock Optimized pre-processing for discrimination prevention.
\newblock In \emph{Advances in Neural Information Processing Systems}, pages
  3992--4001.

\bibitem[{Pleiss et~al.(2017)Pleiss, Raghavan, Wu, Kleinberg, and
  Weinberger}]{Pleiss:17}
Geoff Pleiss, Manish Raghavan, Felix Wu, Jon~M. Kleinberg, and Kilian~Q.
  Weinberger. 2017.
\newblock On fairness and calibration.
\newblock In \emph{Advances in Neural Information Processing Systems}, pages
  5680--5689.

\bibitem[{Ravfogel et~al.(2020)Ravfogel, Elazar, Gonen, Twiton, and
  Goldberg}]{Shauli:20}
Shauli Ravfogel, Yanai Elazar, Hila Gonen, Michael Twiton, and Yoav Goldberg.
  2020.
\newblock Null it out: {G}uarding protected attributes by iterative nullspace
  projection.
\newblock In \emph{Proceedings of the 58th Annual Meeting of the Association
  for Computational Linguistics}, pages 7237--7256.

\bibitem[{Roh et~al.(2020)Roh, Lee, Whang, and Suh}]{Roh:20}
Yuji Roh, Kangwook Lee, Steven Whang, and Changho Suh. 2020.
\newblock Fr-train: {A} mutual information-based approach to fair and robust
  training.
\newblock In \emph{Proceedings of the 37th International Conference on Machine
  Learning}, pages 8147--8157.

\bibitem[{Roh et~al.(2021)Roh, Lee, Whang, and Suh}]{Roh:21}
Yuji Roh, Kangwook Lee, Steven~Euijong Whang, and Changho Suh. 2021.
\newblock Fairbatch: {B}atch selection for model fairness.
\newblock In \emph{Proceedings of the 9th International Conference on Learning
  Representations}.

\bibitem[{Sharifi{-}Malvajerdi et~al.(2019)Sharifi{-}Malvajerdi, Kearns, and
  Roth}]{Saeed:19}
Saeed Sharifi{-}Malvajerdi, Michael~J. Kearns, and Aaron Roth. 2019.
\newblock Average individual fairness: {A}lgorithms, generalization and
  experiments.
\newblock In \emph{Advances in Neural Information Processing Systems}, pages
  8240--8249.

\bibitem[{Shen et~al.(2021)Shen, Han, Cohn, Baldwin, and Frermann}]{Shen:21}
Aili Shen, Xudong Han, Trevor Cohn, Timothy Baldwin, and Lea Frermann. 2021.
\newblock Contrastive learning for fair representations.
\newblock \emph{arXiv preprint arXiv:2109.10645}.

\bibitem[{Subramanian et~al.(2021{\natexlab{a}})Subramanian, Han, Baldwin,
  Cohn, and Frermann}]{Subramanian:21b}
Shivashankar Subramanian, Xudong Han, Timothy Baldwin, Trevor Cohn, and Lea
  Frermann. 2021{\natexlab{a}}.
\newblock Evaluating debiasing techniques for intersectional biases.
\newblock In \emph{Proceedings of the 2021 Conference on Empirical Methods in
  Natural Language Processing}, pages 2492--2498.

\bibitem[{Subramanian et~al.(2021{\natexlab{b}})Subramanian, Rahimi, Baldwin,
  Cohn, and Frermann}]{Subramanian:21}
Shivashankar Subramanian, Afshin Rahimi, Timothy Baldwin, Trevor Cohn, and Lea
  Frermann. 2021{\natexlab{b}}.
\newblock Fairness-aware class imbalanced learning.
\newblock In \emph{Proceedings of the 2021 Conference on Empirical Methods in
  Natural Language Processing}, pages 2045--2051.

\bibitem[{Swayamdipta et~al.(2020)Swayamdipta, Schwartz, Lourie, Wang,
  Hajishirzi, Smith, and Choi}]{Swayamdipta:20}
Swabha Swayamdipta, Roy Schwartz, Nicholas Lourie, Yizhong Wang, Hannaneh
  Hajishirzi, Noah~A. Smith, and Yejin Choi. 2020.
\newblock Dataset cartography: {M}apping and diagnosing datasets with training
  dynamics.
\newblock In \emph{Proceedings of the 2020 Conference on Empirical Methods in
  Natural Language Processing}, pages 9275--9293.

\bibitem[{Wallace et~al.(2011)Wallace, Small, Brodley, and
  Trikalinos}]{Wallace:11}
Byron~C. Wallace, Kevin Small, Carla~E. Brodley, and Thomas~A. Trikalinos.
  2011.
\newblock Class imbalance, redux.
\newblock In \emph{Proceedings of the 11th {IEEE} International Conference on
  Data Mining}, pages 754--763.

\bibitem[{Wang et~al.(2019)Wang, Zhao, Yatskar, Chang, and Ordonez}]{Wang:19}
Tianlu Wang, Jieyu Zhao, Mark Yatskar, Kai-Wei Chang, and Vicente Ordonez.
  2019.
\newblock Balanced datasets are not enough: {E}stimating and mitigating gender
  bias in deep image representations.
\newblock In \emph{Proceedings of the IEEE/CVF International Conference on
  Computer Vision}, pages 5310--5319.

\bibitem[{Wu et~al.(2019)Wu, Zhang, and Wu}]{Wu:19}
Yongkai Wu, Lu~Zhang, and Xintao Wu. 2019.
\newblock Counterfactual fairness: {U}nidentification, bound and algorithm.
\newblock In \emph{Proceedings of the Twenty-Eighth International Joint
  Conference on Artificial Intelligence}, pages 1438--1444.

\bibitem[{Xu et~al.(2018)Xu, Yuan, Zhang, and Wu}]{Xu:18}
Depeng Xu, Shuhan Yuan, Lu~Zhang, and Xintao Wu. 2018.
\newblock Fairgan: {F}airness-aware generative adversarial networks.
\newblock In \emph{{IEEE} International Conference on Big Data}, pages
  570--575.

\bibitem[{Yurochkin et~al.(2020)Yurochkin, Bower, and Sun}]{Yurochkin:20}
Mikhail Yurochkin, Amanda Bower, and Yuekai Sun. 2020.
\newblock Training individually fair {ML} models with sensitive subspace
  robustness.
\newblock In \emph{Proceedings of the 8th International Conference on Learning
  Representations}.

\bibitem[{Zafar et~al.(2017{\natexlab{a}})Zafar, Valera, Gomez{-}Rodriguez, and
  Gummadi}]{Zafar:17}
Muhammad~Bilal Zafar, Isabel Valera, Manuel Gomez{-}Rodriguez, and Krishna~P.
  Gummadi. 2017{\natexlab{a}}.
\newblock Fairness beyond disparate treatment {\&} disparate impact: {L}earning
  classification without disparate mistreatment.
\newblock In \emph{Proceedings of the 26th International Conference on World
  Wide Web}, pages 1171--1180.

\bibitem[{Zafar et~al.(2017{\natexlab{b}})Zafar, Valera, Gomez{-}Rodriguez, and
  Gummadi}]{Zafar:17b}
Muhammad~Bilal Zafar, Isabel Valera, Manuel Gomez{-}Rodriguez, and Krishna~P.
  Gummadi. 2017{\natexlab{b}}.
\newblock Fairness constraints: {M}echanisms for fair classification.
\newblock In \emph{Proceedings of the 20th International Conference on
  Artificial Intelligence and Statistics}, pages 962--970.

\bibitem[{Zhang et~al.(2018)Zhang, Lemoine, and Mitchell}]{Zhanghu:18}
Brian~Hu Zhang, Blake Lemoine, and Margaret Mitchell. 2018.
\newblock Mitigating unwanted biases with adversarial learning.
\newblock In \emph{Proceedings of the 2018 {AAAI/ACM} Conference on AI, Ethics,
  and Society}, pages 335--340.

\bibitem[{Zhang and Bareinboim(2018{\natexlab{a}})}]{Zhang:18}
Junzhe Zhang and Elias Bareinboim. 2018{\natexlab{a}}.
\newblock Equality of opportunity in classification: {A} causal approach.
\newblock In \emph{Advances in Neural Information Processing Systems}, pages
  3675--3685.

\bibitem[{Zhang and Bareinboim(2018{\natexlab{b}})}]{Zhang:18b}
Junzhe Zhang and Elias Bareinboim. 2018{\natexlab{b}}.
\newblock Fairness in decision-making - the causal explanation formula.
\newblock In \emph{Proceedings of the Thirty-Second {AAAI} Conference on
  Artificial Intelligence}, pages 2037--2045.

\bibitem[{Zhao et~al.(2020)Zhao, Coston, Adel, and Gordon}]{Zhao:20}
Han Zhao, Amanda Coston, Tameem Adel, and Geoffrey~J. Gordon. 2020.
\newblock Conditional learning of fair representations.
\newblock In \emph{Proceedings of the 8th International Conference on Learning
  Representations}.

\bibitem[{Zhao et~al.(2017)Zhao, Wang, Yatskar, Ordonez, and Chang}]{Zhao:17}
Jieyu Zhao, Tianlu Wang, Mark Yatskar, Vicente Ordonez, and Kai-Wei Chang.
  2017.
\newblock Men also like shopping: {R}educing gender bias amplification using
  corpus-level constraints.
\newblock In \emph{Proceedings of the 2017 Conference on Empirical Methods in
  Natural Language Processing}, pages 2979--2989.

\end{thebibliography}

\newpage
\appendix

\section{Experimental Settings}

\subsection{\diffadv Setup}
For \diffadv, we use 3 sub-discriminators as \citet{Han:21}, where each sub-discriminator consists of two MLP layers with a hidden size of $256$, followed by a classifier layer to predict the protected attribute. Sub-discriminators are optimised for
at most $100$ epochs after each epoch of main model training, leading to extra training time.

\subsection{Hyperparameter Settings for Twitter Sentiment Analysis}
\label{hyperparameter_moji}

For all models except for \diffadv, the learning rate is $3e-3$, and the batch size is 2,048. For \inlp, following \citet{Shauli:20}, we use $300$ linear SVM classifiers. For \diffadv, the learning rate is $1e-3$, and the batch size is 2,048, the number of discriminators is $3$, $\lambda_{\text{adv}}$ is $0.5$, and $\lambda_{\text{diff}}$ is $1e-3$. For \fairbatch, $\alpha$ is set as $0.1$. For both \difference and \mean, $\lambda$ is set as $0.5$. All hyperparameters are finetuned on the \moji dev set.

\subsection{Hyperparameter Settings for Profession Classification}
\label{hyperparameter_bios}

For all models except for \diffadv, the learning rate is $3e-3$, and the batch size is 2,048. For \inlp, following \citet{Shauli:20}, we use $300$ linear SVM classifiers. For \diffadv, the learning rate is $1e-2$, and the batch size is 1,024, the number of discriminators is $3$, $\lambda_{\text{adv}}$ is $1e-2$, and $\lambda_{\text{diff}}$ is $1e4$. For \fairbatch, $\alpha$ is set as $5e-2$. For \difference, $\lambda$ is set as $1e-2$, and for \mean, $\lambda$ is set as $5e-3$. All hyperparameters are finetuned on the \bios dev set.


\section{Analysis}

\subsection{Ablation Study hyperparameter Settings}
\label{ablation_parameters}
For all models, we have tuned the hyperparameter $\lambda$ and selected model based on performance on the dev set. On the \moji dataset, for \differencemax, $\lambda=2$, for \differencemin, $\lambda=0.4$, for \meanmax, $\lambda=2$, for \meanmin, $\lambda=0.2$. On the \bios dataset, for \differencemax, $\lambda=0.05$, for \differencemin, $\lambda=0.005$, for \meanmax, $\lambda=0.005$, for \meanmin, $\lambda=1e-4$. 


\end{document}